\documentclass[runningheads]{llncs}

\usepackage{eccv}

\usepackage{tabularx}
\usepackage{eccvabbrv}

\usepackage{graphicx}
\usepackage{booktabs}

\usepackage[accsupp]{axessibility}  

\usepackage{hyperref}

\usepackage{orcidlink}

\begin{document}

\title{EMO: Emote Portrait Alive - Generating Expressive Portrait Videos with Audio2Video Diffusion Model under Weak Conditions}

\titlerunning{EMO-Emote Portrait Alive}

\author{Linrui Tian\orcidlink{0000-0003-1202-6040} \and
Qi Wang \and
Bang Zhang \and
Liefeng Bo}

\authorrunning{L. Tian, Q. Wang, B. Zhang, and L. Bo}

\institute{Institute for Intelligent Computing, Alibaba Group\\
\email{\{tianlinrui.tlr, wilson.wq, zhangbang.zb, liefeng.bo\}@alibaba-inc.com}\\
\url{https://humanaigc.github.io/emote-portrait-alive/}}

\maketitle

\begin{figure}[h]
  \centering
  \includegraphics[width=0.8\textwidth]{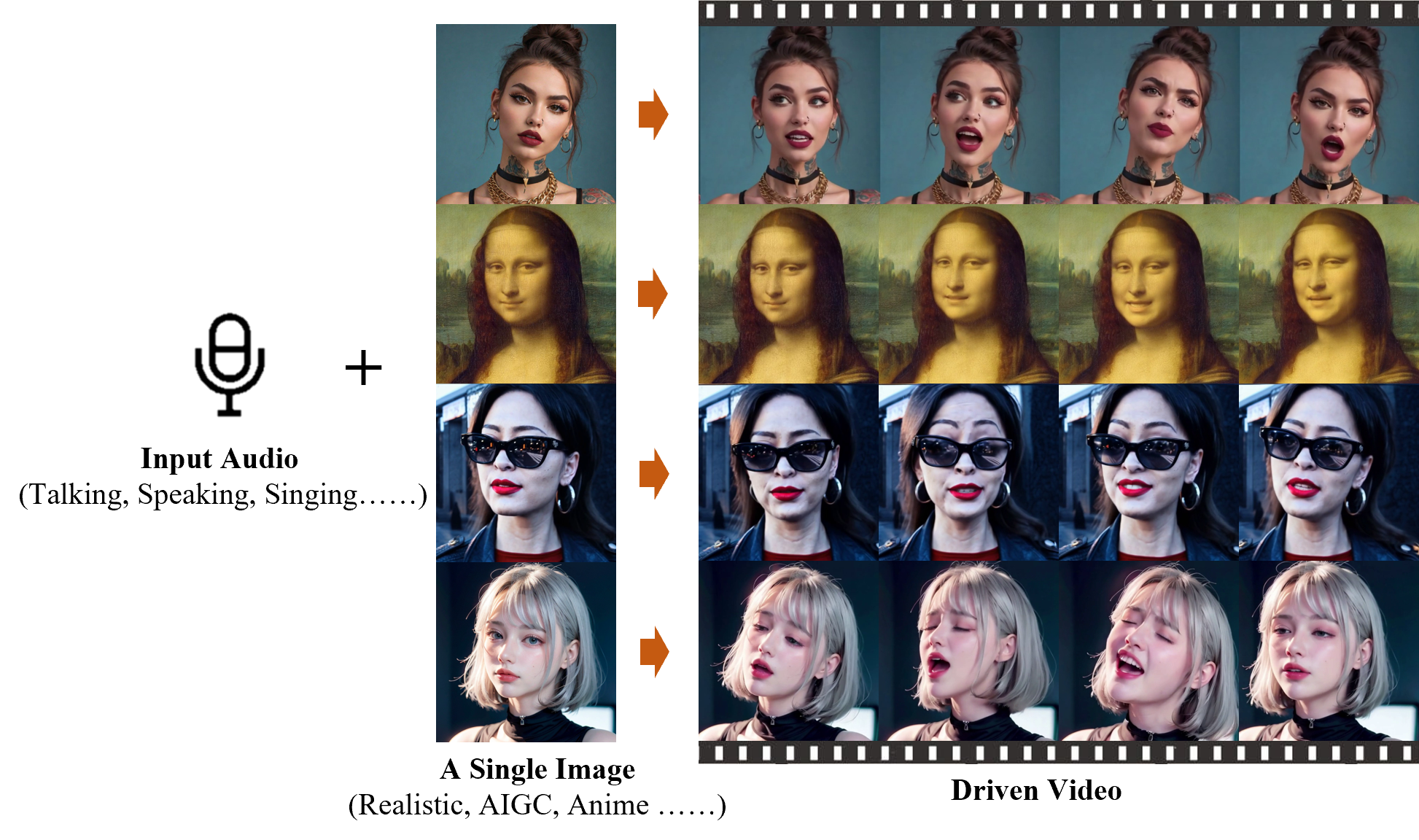}
  \caption{We proposed EMO, an expressive audio-driven portrait-video generation framework. Input a \textbf{single reference image} and the vocal audio, e.g. talking and singing, our method can generate vocal avatar videos with expressive facial expressions, and various head poses, meanwhile, we can generate videos with \textbf{any duration} depending on the length of input audio.
  }
  \label{fig:intro}
\end{figure}

\begin{abstract}
In this work, we tackle the challenge of enhancing the realism and expressiveness in talking head video generation by focusing on the dynamic and nuanced relationship between audio cues and facial movements. We identify the limitations of traditional techniques that often fail to capture the full spectrum of human expressions and the uniqueness of individual facial styles. To address these issues, we propose EMO, a novel framework that utilizes a direct audio-to-video synthesis approach, bypassing the need for intermediate 3D models or facial landmarks. Our method ensures seamless frame transitions and consistent identity preservation throughout the video, resulting in highly expressive and lifelike animations. Experimental results demonstrate that EMO is able to produce not only convincing speaking videos but also singing videos in various styles, significantly outperforming existing state-of-the-art methodologies in terms of expressiveness and realism.

  \keywords{Diffusion Models \and Video Generation \and Talking Head}
\end{abstract}


\section{Introduction}
\label{sec:intro}
Diffusion Models have revolutionized the landscape of generative models, showcasing unparalleled capabilities in generating high-fidelity images\cite{ddpm,ldm,DiffusionNonequ,DiffusionBeatGAN,DiT}. These advancements have extended into video generation, sparking interest in leveraging these models for dynamic and engaging visual storytelling\cite{ho2022video,bartal2024lumiere,2023i2vgenxl}. Beyond the general video generation, human-centric video generation, including portrait video generation\cite{sun2023vividtalk, diffused_head, sadtalker, dreamtalk} and human animation\cite{animatediff, animate_anyone}, has attracted considerable attention for its utility in creating digital human avatars and enhancing film production. 
A notably prominent area within this field is "talking head" video generation, which aims to synthesize 
a head video that matches with the input audio, while capturing the facial expressions, head movements and lip motion. 

However, translating audio into head animation, such as facial expressions or head movements, is challenging due to the ambiguous and one-to-many mapping relationship. 
Most research on talking heads divides the process into head motion and facial expression components\cite{sun2023vividtalk}. For head motion, some talking head techniques struggle to manage this aspect effectively and often resort to using a predefined pose sequence from an existing video\cite{wav2lip,Diff2Lip,syncnet} or use separate networks to individually address head poses and facial expressions\cite{sun2023vividtalk}. In terms of facial expressions, some approaches opt for explicit intermediate signals like 3D face models or 2D facial landmarks to guide the generation\cite{sun2023vividtalk,sadtalker}. 
While these methods enhance the fidelity of specific aspects like lip synchronization, they tend to restrict the overall expressiveness and naturalness of the generated content. For example, the subtleties of spontaneous gestures or the nuanced expressions linked to speech's emotional tone are often lost in translation, leading to less lifelike results.
Therefore, to create highly expressive videos of talking head, it's crucial to move away from the constraints of strong prior information and fully leverage the generative potential of the model.
In this paper, we introduce EMO (Emote Portrait Alive), a novel talking head framework that transforms a single portrait image into an expressive video, synchronized with audio, without relying on intermediate 3D representations or predefined motion templates. 
EMO harnesses the generative power of Diffusion Models to directly capture the intricate audio-visual correlations, facilitating the generation of dynamic and lifelike talking head videos.
Specifically, our method extends Stable Diffusion for video by incorporating temporal modules and 3D convolutions, see Sec.~\ref{sec:backbone}. 
In order to lear the correlation between audio and videos, we introduce an audio feature extractor and employ attention module to modulate audio features into the backbone, see Sec.~\ref{sec:audiolayer}.
To ensure stability without compromising expressiveness, we introduce novel mechanisms like the Face Locator and Speed Layers, to perform as week conditions for guiding the general area of the target face and the approximate speed level of movements. 
Unlike the strong priors, such as 3D intermediate representation, used by previous works, these conditions are not able to strictly constrains the face position and the speed of the generated videos, and therefore do not diminish the expressiveness of the generated videos (for more details, see Sec.~\ref{sec:ablation}). Additionally, we introduced Reference Net\cite{animate_anyone} to ensure consistent facial identity throughout the video (see Sec.~\ref{sec:referencenet}) and implemented Motion Frame module to maintain continuity between adjacent video clips, enabling the generation of seamlessly infinite videos, Sec.~\ref{sec:temporalmodule}.

To train our model, we constructed a vast and diverse audio-video dataset, amassing over 250 hours of footage. This expansive dataset encompasses a wide range of content, including speeches, film and television clips, and singing performances, and covers multiple languages such as Chinese and English. The rich variety of speaking and singing videos ensures that our training material captures a broad spectrum of human expressions and vocal styles, providing a solid foundation for the development of EMO. 
We conducted extensive experiments across multiple datasets, comparing our model with various state-of-the-art methods. Our model outperforms these on several metrics, including a new metric we introduced, E-FID (Expression-FID), designed to evaluate the expressiveness of generated videos, thereby demonstrating superior performance.
In summary, we make the following contributions: 1) We propose a novel talking head framework based on fully generative model without any 3D intermediates or pose prior. 2) Our method outperforms other existing SOTA methods in terms of expressiveness and realism of the talking head video. 3) We use weak conditions to stabilize the generation process without incurring significant loss in performance.

\section{Related Work}
\textbf{Diffusion Models}
Diffusion Models have demonstrated remarkable capabilities across various domains, including image synthesis\cite{DiffusionBeatGAN,ddpm}, image editing\cite{imageedit-imagic, dragdiffusion}, video generation\cite{animatediff, animate_anyone} and even 3D  content generation\cite{dreamfusion, magic3d}. Among them, Stable Diffusion (SD)\cite{ldm} stands out as a representative example, employing a UNet architecture to iteratively generate images with notable text-to-image capabilities, following extensive training on large text-image datasets\cite{laion5b}. 
These pretrained models have found widespread application in a variety of image and video generation tasks\cite{animate_anyone,animatediff}. 
Additionally, some recent works adopt a DiT (Diffusion-in-Transformer)\cite{DiT} which alters the UNet with a Transformer incpororating temporal modules and 3D Convoluations, enabling support for larger-scale data and model parameters. By training the entire text-to-video model from scratch, it achieves superior video generation results\cite{latte}. Also, some efforts have delved into applying Diffusion Models for talking head generation, producing promising results that highlight the capability of these models in crafting realistic talking head videos\cite{diffused_head,dreamtalk}. 

\textbf{Audio-driven talking head generation}
Audio-driven talking head generation can be broadly catgorized into two approaches:video-based methods\cite{wen2020audiodvp,faceformer2022, wav2lip, shen2023difftalk, guan2023stylesync} and single-image \cite{sadtalker,sun2023vividtalk,dreamtalk, liu2023moda}.
video-based talking head generation allows for direct editing on an input video segment. For example, Wav2Lip\cite{wav2lip} regenerates lip movements in a video based on audio, using a discriminator for audio-lip sync. Its limitation is relying on a base video, leading to fixed head movements and only generating mouth movements, which can limit realism.
For single-image talking head generation, a reference photo is utilized to generate a video that mirrors the appearance of the photo. 
\cite{sun2023vividtalk} proposes to generate the head motion and facial expressions independently by learning blendshapes and head poses. These are then used to create a 3D facial mesh, serving as an intermediate representation to guide the final video frame generation. Similarly, \cite{sadtalker} employs a 3D Morphable Model (3DMM) as an intermediate representation for generating talking head video.
A common issue with these methods is the limited representational capacity of the 3D mesh, which constrains the overall expressiveness and realism of the generated videos. Additionally, both methods are based on non-diffusion models, which further limits the performance of the generated results. \cite{dreamtalk} attempts to use diffusion models for talking head generation, but instead of applying directly to image frames, it employs them to generate coefficients for 3DMM. Compared to the previous two methods, Dreamtalk offers some improvement in the results, but it still falls short of achieving highly natural facial video generation.

\section{Method}
\label{sec:method}
Given a single reference image of a character portrait, our approach can generate a video synchronized with an input voice audio clip, preserving the natural head motion and vivid expression in harmony with the tonal variances of the provided vocal audio. By creating a seamless series of cascaded video, our model facilitates the generation of long-duration talking portrait videos with consistent identity and coherent motion, which are crucial for realistic applications.

\subsection{Preliminaries}

Our methodology employs Stable Diffusion (SD) as the foundational framework. SD is a widely-utilized text-to-image (T2I) model that evolves from the Latent Diffusion Model (LDM)\cite{ldm}. It utilizes an autoencoder Variational Autoencoder (VAE)\cite{vae} to map the original image feature distribution $x_0$ into latent space $z_0$, encoding the image as $z_0=\mathbf{E}(x_0)$ and reconstructing the latent features as $x_0=\mathbf{D}(z_0)$. This architecture offers the advantage of reducing computational costs while maintaining high visual fidelity. Based on the Denoising Diffusion Probabilistic Model (DDPM)\cite{ddpm} or the Denoising Diffusion Implicit Model (DDIM)\cite{ddim} approach, SD introduces Gaussian noise $\epsilon$ to the latent $z_0$ to produce a noisy latent $z_t$ at a specific timestep $t$. During inference, SD aims to remove the noise $\epsilon$ from the latent $z_t$ and incorporates text control to achieve the desired outcome by integrating text features. The training objective for this denoising process is expressed as:
\begin{equation}
\mathcal{L}=\mathbb{E}_{t,c, z_t, \epsilon}\left[ ||\epsilon - \epsilon_{\theta}(z_t, t, c)||^2\right]
\end{equation}
where $c$ represents the text features, which are obtained from the token prompt via the CLIP\cite{clip} ViT-L/14 text encoder. In SD, $\epsilon_{\theta}$ is realized through a modified UNet\cite{unet} model, which employs the cross-attention mechanism to fuse $c$ with the latent features.

\subsection{Network Pipelines}

\begin{figure}[tb]
  \centering
  \includegraphics[width=\textwidth]{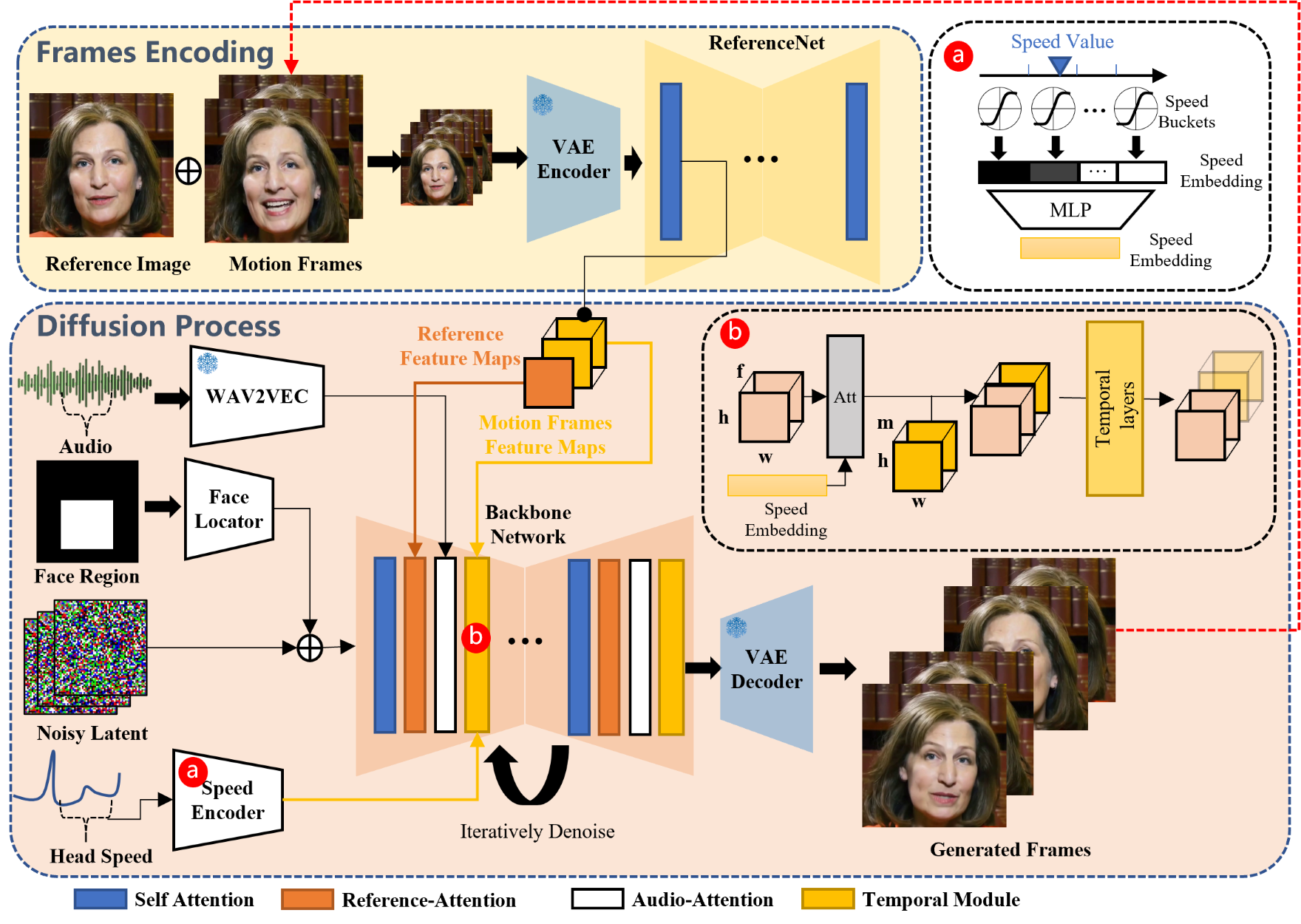}
  \caption{Overview of the proposed method. Our framework is mainly constituted with two stages. In the initial stage, termed Frames Encoding, the ReferenceNet is deployed to extract features from the reference image and motion frames. Subsequently, during the Diffusion Process stage, a pretrained audio encoder processes the audio embedding. The facial region mask is integrated with multi-frame noise to govern the generation of facial imagery. This is followed by the employment of the Backbone Network to facilitate the denoising operation. Within the Backbone Network, two forms of attention mechanisms are applied: Reference-Attention and Audio-Attention. These mechanisms are essential for preserving the character's identity and modulating the character's movements, respectively. Additionally, Temporal Modules are utilized to manipulate the temporal dimension, and adjust the velocity of motion.
  }
  \label{fig:pipeline}
\end{figure}

The overview of our method is shown in Figure \ref{fig:pipeline}. Our \textbf{Backbone Network} get the multi-frame noise latent input, and try to denoise them to the consecutive video frames during each time step, the Backbone Network has the similar UNet structure configuration with the SD 1.5. 1) Similar to previous work, to ensure the continuity between generated frames, the Backbone Network is embedded with \textbf{temporal modules}. 2) To maintain the ID consistency of the portrait in the generated frames, we deploy a UNet structure called \textbf{ReferenceNet} parallel to the Backbone, it input the reference image to get the features. 3) To drive the character speaking motion, \textbf{audio layers} are utilized to encode the voice features. 4) To make the motion of talking character controllable and stable, we use the \textbf{face locator} and \textbf{speed layers} to provide weak conditions.

\subsubsection{Backbone Network.} 
\label{sec:backbone}
In our work, the prompt embedding is not utilized; hence, we have adapted the cross-attention layers in the SD 1.5 UNet structure to reference-attention layers. These modified layers now take reference features from ReferenceNet as input rather than text embeddings.

\subsubsection{Audio Layers.}
\label{sec:audiolayer}
The pronunciation and tone in the voice is the main driven sign to the generated character. The features extracted from the input audio sequence by the various blocks of the pretrained wav2vec\cite{wav2vec} are concatenated to yield the audio representation embedding, $A^{(f)}$, for the $f$th frame. However, the motion might be influenced by the future/past audio segments, for example, opening mouth and inhaling before speaking. To address that, we define voice features of each generated frame by concatenating the features of nearby frames: 
$\mathbf{A}^{(f)}=\oplus\{A^{(f-m)},...A^{(f)},...A^{(f+m)}\}$, $m$  is the number of additional features from one side. To inject the voice features into the generation procedure, we add audio-attention layers performing a cross attention mechanism between the latent code and $\mathbf{A}$ after each ref-attention layers in the Backbone Network. 


\subsubsection{ReferenceNet.} 
\label{sec:referencenet}
The ReferenceNet possesses a structure identical to that of the Backbone Network and serves to extract features from input images. Prior research\cite{googletryon, animate_anyone} has underscored the profound influence of utilizing analogous structures in maintaining the consistency of the target object's identity. In our study, both the ReferenceNet and the Backbone Network inherit weights from the original SD UNet. The reference image is inputted into the ReferenceNet to extract the reference feature maps outputs from the self-attention layers. During the Backbone denoising procedure, the features of corresponding layers undergo a reference-attention layers with the extracted feature maps.

\subsubsection{Temporal Modules.} 
\label{sec:temporalmodule}
Informed by the architectural concepts of AnimateDiff, we apply self-attention temporal layers to the features within frames. Specifically, we reconfigure the input feature map $x \in \mathbb{R}^{b \times c \times f \times h \times w}$ to the shape ${(b \times h \times w) \times f \times c}$. Here, $b$ stands for the batch size, $h$ and $w$ indicate the spatial dimensions of the feature map, $f$ is the count of generated frames, and $c$ is the feature dimension. Notably, we direct the self-attention across the temporal dimension $f$, to effectively capture the dynamic content of the video. The temporal layers are inserted at each resolution stratum of the Backbone Network. 
Most diffusion-based video generation models are inherently limited by their design to produce a predetermined number of frames, thereby constraining the creation of extended video. This limitation is particularly impactful in applications of talking head videos, where a sufficient duration is essential for the articulation of meaningful speaking. Some methodologies employ a frame from the end of the preceding clip as the initial frame of the subsequent generation, aiming to maintain a seamless transition across concatenated segments. Inspired by that, our approach incorporates the last $n$ frames, termed 'motion frames' from the previously generated clip to enhance cross-clip consistency. Specifically, these $n$ motion frames are fed into the ReferenceNet to pre-extract multi-resolution motion feature maps. During the denoising process within the Backbone Network, we merge the temporal layer inputs with the pre-extracted motion features of matching resolution along the frames dimension. This straightforward method effectively ensures coherence among various clips. For the generation of the first video clip, we initialize the motion frames as zero maps.

It should be noted that while the Backbone Network may be iterated multiple times to denoise the noisy frames, the target image and motion frames are concatenated and input into the ReferenceNet only once. Consequently, the extracted features are reused throughout the process, ensuring that there is no substantial increase in computational time during inference.

\subsubsection{Face Locator and Speed Layers.} Temporal modules can guarantee continuity of the generated frames and seamless transitions between video clips, however, they are insufficient to ensure the consistency and stability of the generated character's motion across the clips, due to the independent generation process. Previous works use some signal to control the character motion, e.g. skeleton\cite{animate_anyone}, blendshape\cite{sadtalker}, or 3DMM\cite{sun2023vividtalk}, nevertheless, employing these control signals may be not good in creating lively facial expressions and actions due to their limited degrees of freedom, and the inadequate labeling during training stage are insufficient to capture the full range of facial dynamics. Additionally, the same control signals could result in discrepancies between different characters, failing to account for individual nuances. Enabling the generation of control signals may be a viable approach\cite{sun2023vividtalk}, yet producing lifelike motion remains a challenge. Therefore, we opt for a "weak" control signal approach. 

Specifically, as shown in Figure \ref{fig:pipeline}, we utilize a mask $\mathbf{M}=\bigcup_{i=1}^{f} M^{i}$ as the face region, which encompasses the facial bounding box (bbox) regions of the video clip. We employ the Face Locator, which consists of lightweight convolutional layers designed to encode the bounding box mask. The resulting encoded mask is then added to the noisy latent representation before being fed into the Backbone. We can use the mask to control where the character face should be generated. 


However, creating consistent and smooth motion between clips is challenging due to variations in head motion frequency during separate generation processes. To address this issue, we incorporate the target head motion speed into the generation. More precisely, we consider the head rotation velocity $w^f$ in frame $f$ and divide the range of speeds into $d$ discrete speed buckets, each representing a different velocity level. Each bucket has a central value $c_i\in \{c_1,...,c_d\}$ and a radius $r_i \in \{r_1, ..., r_d\}$. We retarget $w^f$ to a vector $\mathbf{s}\in \mathbb{R}^d$, where the $i$th value notated by $s_i = \tanh((w^f-c_i)/r_i*3)$.
Similar to the method used in the audio layers, the head rotation speed embedding for each frame is given by $\mathbf{S}^{f}=\oplus\{\mathbf{s}^{(f-m)},\ldots,\mathbf{s}^{(f)},\ldots,\mathbf{s}^{(f+m)}\}$. The speed embedding of each clip denoted by $\mathbf{S}\in\mathbb{R}^{b\times f \times (2m+1)d}$ is subsequently processed by an MLP into a speed feature map $\mathbf{F}\in\mathbb{R}^{b\times f\times l}$. Within the temporal layers, we repeat $\mathbf{F}$ to the shape ${(b \times h \times w) \times f \times l}$ and implement a cross-attention mechanism that operates between the speed features and the reshaped feature map across the temporal dimension $f$. By doing so and specifying a target speed, we can synchronize the rotation speed and frequency of the generated character's head across different clips. Combined with the facial position control provided by the Face Locator, the resulting output can be both stable and controllable.

It should also be noted that the specified face region and assigned speed does not constitute strong control conditions. In the context of face locator, since the $\mathbf{M}$ is the union area of the entire video clip, indicating a sizeable region within which facial movement is permissible, thereby ensuring that the head is not restricted to a static posture. With regard to the speed layers, the difficulty in accurately estimating human head rotation speed for dataset labeling means that the predicted speed sequence is inherently noisy. Consequently, the generated head motion can only approximate the designated speed level. This limitation motivates the design of our speed buckets framework.

\subsection{Training Strategies}

The training process is structured into three stages. The first stage is the image pretraining, where the Backbone Network, the ReferenceNet, and the Face Locator are token into training, in this stage, the Backbone takes a single frame as input, while ReferenceNet handles a distinct, randomly chosen frame from the same video clip. Both the Backbone and the ReferenceNet initialize weights from the original SD. In the second stage, we introduce the video training, where the temporal modules and the audio layers are incorporated, $n + f$ contiguous frames are sampled from the video clip, with the started $n$ frames are motion frames. The temporal modules initialize weights from AnimateDiff\cite{animatediff}. In the last stage, the speed layers are integrated, we only train the temporal modules and the speed layers in this stage. This strategic decision deliberately omits the audio layers from the training process. Because the speaking character's expression, mouth motion, and the frequency of the head movement, are predominantly influenced by the audio. Consequently, there appears to be a correlation between these elements, the model might be prompted to drive the character's motion based on the speed signal rather than the audio. Our experimental results suggest that simultaneous training of both the speed and audio layers undermines the driven ability of the audio on the character's motions.

\section{Experiments}

\subsection{Implementations}
We collected approximately 250 hours of talking head videos from the internet and supplemented this with the HDTF\cite{hdtf} and VFHQ\cite{vfhq} datasets to train our models. As the VFHQ dataset lacks audio, it is only used in the first training stage. We apply the MediaPipe\cite{mediapipe} to obtain the facial bbox. Head rotation velocity was labeled by extracting the 6-DoF head pose for each frame using facial landmarks, followed by calculating the rotational degrees between frames.

The video clips sampled from the dataset are resized and cropped to $512\times{512}$. In the first training stage, the reference image and the target frame are sampled from the video clip separately, we trained the Backbone Network and the ReferneceNet with a batch size of 48. In the second and the third stage, we set $f=12$ as the generated video length, and the motion frames number is set to $n=4$, we adopt a bath size of 4 for training. The additional features number $m$ is set to 2. The learning rate for all stages are set to 1e-5. During the inference, we use the sampling algorithm of DDIM to generate the video clip for 40 steps, we assign a constant speed value for each frame generation. The time consumption of our method is about 15 seconds for one batch ($f=12$ frames).

\subsection{Experiments Setup}

For methods comparisons, we partitioned the HDTF dataset, allocating 10\% as the test set and reserving the remaining 90\% for training. We took precautions to ensure that there was no overlap of character IDs between these two subsets. Additionally, to evaluate the methods in more variable scenarios, 1k video clips were extracted from the our collected internet video dataset, each with a duration of approximately 4 seconds. These clips predominantly feature expressive portrait videos with a substantial proportion depicting singing activities. Compared to the HDTF, the video dataset exhibits a broader diversity in terms of facial expressions, and the range of head motions.

We compare our methods with some previous works including: Wav2Lip\cite{wav2lip}, SadTalker\cite{sadtalker}, DreamTalk\cite{dreamtalk}, MakeItTalk\cite{MakeItTalk}. Additionally, we generated results using the released code from Diffused Heads\cite{diffused_head}, however, the model is trained on CREMA\cite{crema} dataset which contains only green background, the generated results are suboptimal. Furthermore, the results were compromised by error accumulation across the generated frames. Therefore, we only conduct qualitative comparison with the Diffused Heads approach. For DreamTalk, we utilize the talking style parameters as prescribed by the original authors.

To demonstrate the superiority of the proposed method, we evaluate the model with several metrics. We utilize Fréchet Inception Distance (FID)\cite{fid} to assess the quality of the generated frame\cite{portrait3d}. Additionally, to gauge the preservation of identity in our results, we computed the facial similarity (F-SIM) by extracting and comparing facial features between the generated frames and the reference image. It is important to note that using a single, unvarying reference image could result in deceptively perfect F-SIM scores. Certain methods\cite{wav2lip} might produce only the mouth regions, leaving the rest of the frame identical to the reference image, which could skew results. Therefore, we treat F-SIM as population-reference indices\cite{diffused_head}, with closer approximations to the corresponding ground truth (GT) values indicating better performance. We further employed the Fréchet Video Distance (FVD)\cite{fvd} for the video-level evaluation. The SyncNet\cite{syncnet} score was used to assess the lip synchronization quality, a critical aspect for talking head applications. To evaluate the expressiveness of the facial expressions in the generated videos, we introduce the use of the Expression-FID (E-FID) metric. This involves extracting expression parameters via face reconstruction techniques, as described in \cite{deep3d}. Subsequently, we compute the FID of these expression parameters to quantitatively measure the divergence between the expressions in the synthesized videos and those in the ground truth dataset.

\subsection{Qualitative Comparisons}

\begin{figure}[tbh]
  \centering
  \includegraphics[width=\textwidth]{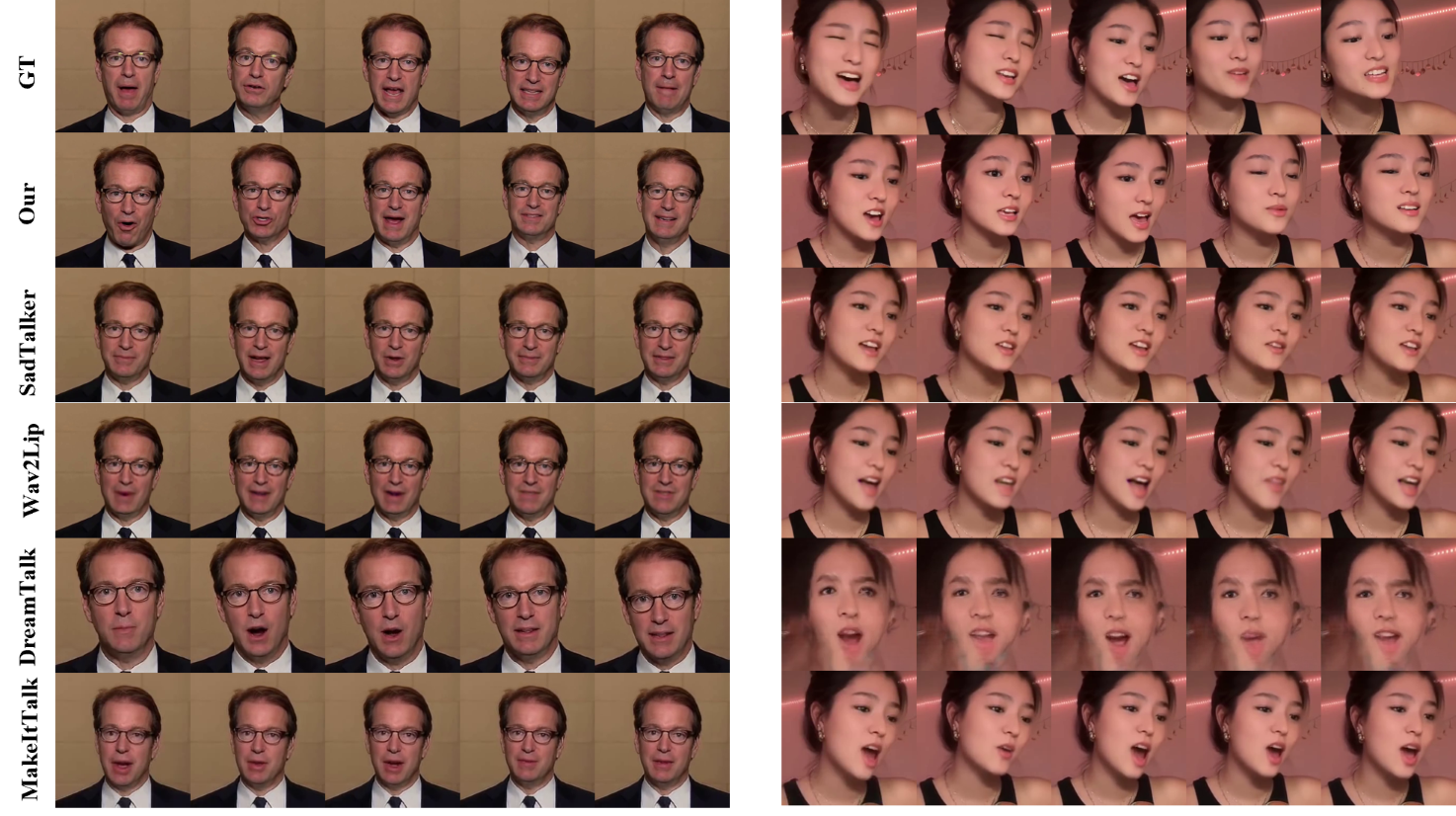}
  \caption{The qualitative comparisons with several talking head generation works. The left column is the generated results on the HDTF dataset. the right column is the results on the internet data.
  }
  \label{fig:quality_com}
\end{figure}

Figure \ref{fig:quality_com} demonstrates the visual results of our method alongside those of earlier approaches. It is observable that Wav2Lip typically synthesizes blurry mouth regions and produces videos characterized by a static head pose and minimal eye movement when a single reference image is provided as input. In the case of DreamTalk\cite{dreamtalk}, the style clips supplied by the authors could distort the original faces, also constrain the facial expressions and the dynamism of head movements. In contrast to SadTalker and DreamTalk, our proposed method is capable of generating a greater range of head movements and more dynamic facial expressions. Since we do not utilize direct signal, e.g. blendshape or 3DMM, to control the character motion, these motions are directly driven by the audio, which will be discussed in detail in the following showcases.

\begin{figure}[tb]
  \centering
  \includegraphics[width=\textwidth]{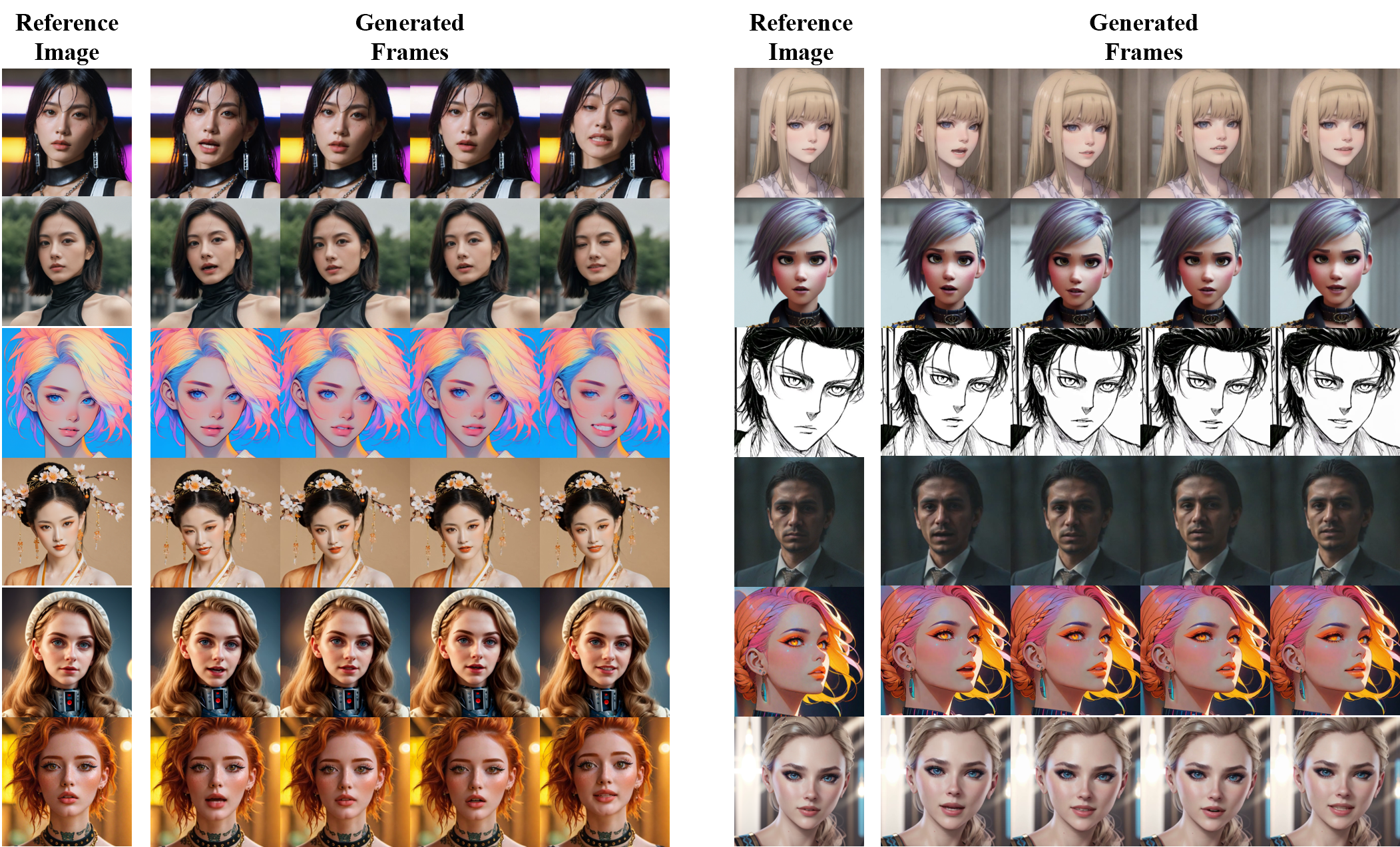}
  \caption{The qualitative results of our method based on \textbf{different portrait styles}. Here we demonstrate 14 generated video clips, in which the characters are driven by the same vocal audio clip. The duration of each generated clip is approximately 8 seconds. Due to space limitations, we only sample four frames from each clip.
  }
  \label{fig:quality_wild_com}
\end{figure}

We further explore the generation of talking head videos across various portrait styles. As illustrated in Figure \ref{fig:quality_wild_com}, the reference images, sourced from Civitai, are synthesized by disparate text-to-image (T2I) models, encompassing characters of diverse styles, namely realistic, anime, and 3D. These characters are animated using identical vocal audio inputs, resulting in approximately uniform lip synchronization across the different styles. Although our model is trained only on the realistic videos, it demonstrates proficiency in producing talking head videos for a wide array of portrait types.

\begin{figure}[tb]
  \centering
  \includegraphics[width=\textwidth]{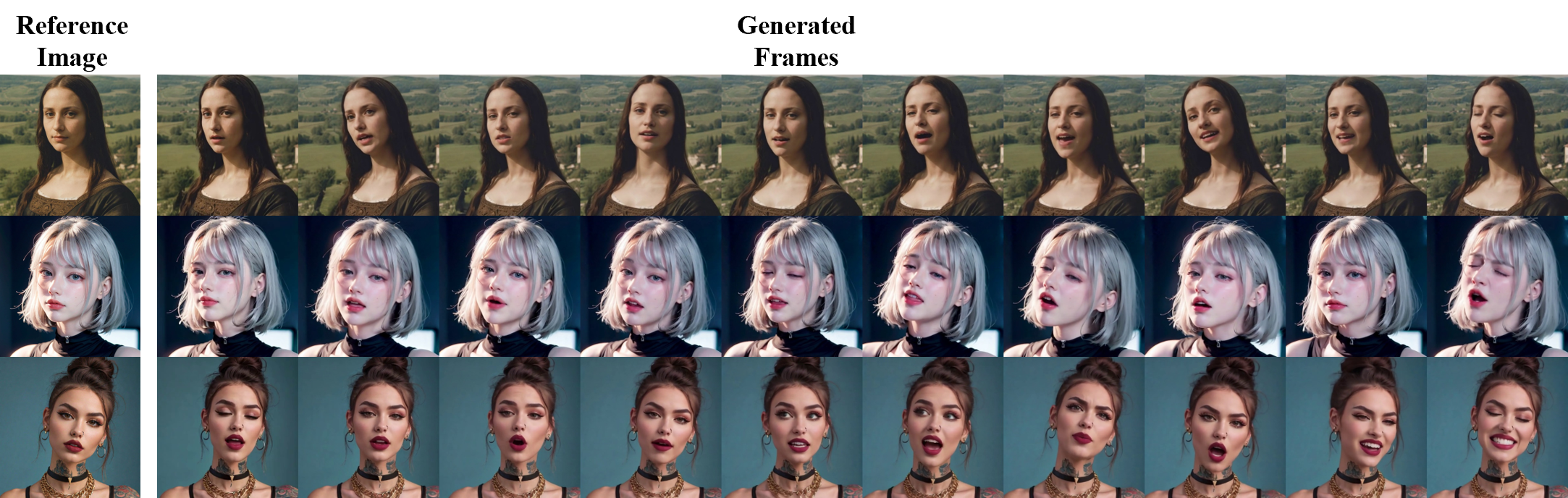}
  \caption{The results generated by our method for voice with a \textbf{strong tonal quality} over a \textbf{prolonged duration}. In each clip, the character is driven by the strong tonal audio, e.g. singing, and the duration of each clip is approximately 1 minute.
  }
  \label{fig:quality_song_com}
\end{figure}

\begin{figure}[tb]
  \centering
  \includegraphics[width=0.6\textwidth]{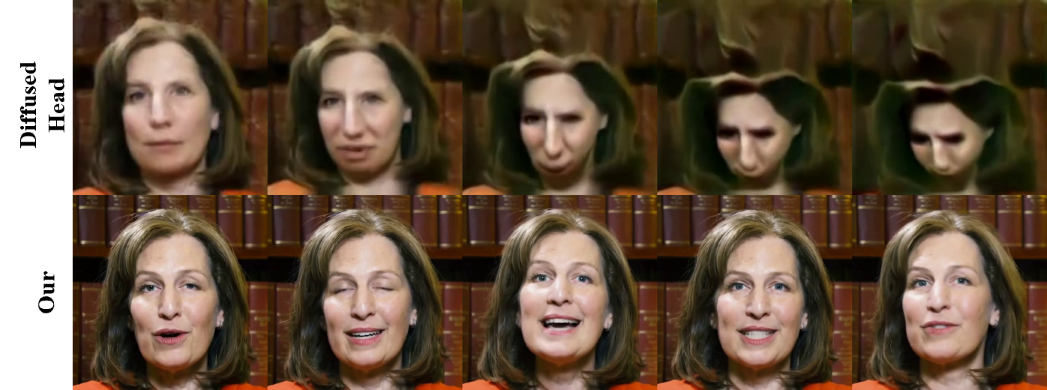}
  \caption{Comparisons with Diffused Heads\cite{diffused_head}, the duration of generated clips is 6 seconds, the results of Diffused Heads have low resolution and are compromised by error accumulation across the generated frames.
  }
  \label{fig:diffused_head_com}
\end{figure}

Figure \ref{fig:quality_song_com} demonstrates that our method is capable of generating richer facial expressions and movements when processing audio with pronounced tonal features. For instance, the examples in the third row reveal that high-pitched vocal tones elicit more intense and animated expressions from the characters.  Moreover, leveraging motion frames allows for the extension of the generated video, we can generate prolonged duration video depending on the length of the input audio. As shown in Figure \ref{fig:quality_song_com} and Figure \ref{fig:diffused_head_com}, our approach preserves the character's identity over extended sequences, even amidst substantial motion.

\subsection{Quantitative Comparisons}


\begin{table}[h!]
  \centering
  \caption{The quantitative comparisons on HDTF and internet data, with several talking head generation works.}
  \label{tab:quantitative_comp}
  \begin{tabularx}{\textwidth}{l>{\centering\arraybackslash}X>{\centering\arraybackslash}X>{\centering\arraybackslash}X>{\centering\arraybackslash}X>{\centering\arraybackslash}X}
    \toprule
    Method & FID$\downarrow$ & SyncNet$\uparrow$ & F-SIM & FVD$\downarrow$ &  E-FID$\downarrow$\\
    \midrule
    Wav2Lip\cite{wav2lip} & {\scriptsize 9.38/31.70} & {\scriptsize \textbf{5.79}/\textbf{4.14}} & {\scriptsize 80.34/78.87} & {\scriptsize 407.93/487.00} & {\scriptsize 0.693/0.652} \\
    SadTalker\cite{sadtalker} & {\scriptsize 10.31/31.37} & {\scriptsize 4.82/2.90} & {\scriptsize 84.56/81.86} & {\scriptsize 214.98/418.19} & {\scriptsize 0.503/0.539} \\
    DreamTalk\cite{dreamtalk} & {\scriptsize 58.80/88.21} & {\scriptsize 3.43/1.29} & {\scriptsize 67.87/56.38} & {\scriptsize 619.05/584.63} & {\scriptsize 2.257/3.548}\\
    MakeItTalk\cite{MakeItTalk} & {\scriptsize 21.73/39.86} & {\scriptsize 2.85/1.64} & {\scriptsize \textbf{76.91}/60.12} & {\scriptsize 350.96/340.55} & {\scriptsize 1.072/0.997} \\
    GT & - & {\scriptsize 7.3/2.69} & {\scriptsize 77.44/72.64} & - & -\\
    
    w/o 250h data & {\scriptsize 10.80/-} & {\scriptsize 5.02/-} & {\scriptsize 79.55/-}  & {\scriptsize 102.78/-} & {\scriptsize 0.215/-} \\
    
    Ours & {\scriptsize \textbf{8.76}/\textbf{17.33}} & {\scriptsize 3.89/2.74} & {\scriptsize 78.96/\textbf{77.16}}  & {\scriptsize \textbf{67.66}/\textbf{192.77}} & {\scriptsize \textbf{0.116}/\textbf{0.187}} \\
    \bottomrule
  \end{tabularx}
\end{table}


Figure \ref{fig:quality_com} illustrates that the internet dataset encompasses a wider variety of facial expressions and an extensive range of head movements, coupled with diverse poses of the reference character. Such variability has the potential to negatively impact performance metrics, as shown in Table \ref{tab:quantitative_comp}.  our results demonstrate a substantial advantage in video quality assessment, as evidenced by the lower FVD scores. Additionally, our method outperforms others in terms of individual frame quality, as indicated by improved FID scores. Wav2Lip has the highest SyncNet confidence score by training with SyncNet as a dicriminator\cite{dreamtalk}, despite not achieving the highest scores on the SyncNet metric, our approach excels in generating lively facial expressions as shown by E-FID. Furthermore, we also investigated the benefit of our module and the impact of the 250 hours dataset by training our model solely on publicly available datasets including VFHQ, HDTF, and CELEB-V \cite{zhu2022celebvhq}. Even in the absence of the 250-hour dataset, our model still exhibits exceptional performance, particularly in terms of FVD and E-FID. While the further improvements on these metrics caused by the collected dataset demonstrates that the extra data could contributes to enhancing video content dynamics and generating a wider variety of expressions.

\subsection{Ablation Studies}
\label{sec:ablation}

\subsubsection{the Impact of the Speed Layer.} 
\label{exp:speed}
The speed layers are designed to ensure the consistency of the head motion frequency between the contiguous generated video clips. During the inference, we assign a constant speed value for every frame. To measure the effect of the speed layers, we generate video clips with different assigned speed value on the HDTF dataset. As shown in Table \ref{tab:speed_layer}, "No Speed" indicates the results generated by the model without the speed layers. "Velocity Variance" represents the average variance of the velocities across individual video sequences, reflecting the consistency of rotational speed within each clip. "Variance of Mean Velocities (VMV)" indicates the variance of the mean velocities across different clips, providing a measure of the variability in head rotation speeds from one clip to the others. Incorporating speed layers significantly enhances the stability of head motion, as evidenced by the reduced "Velocity Variance" and "VMV" in comparison to the baseline "No Speed" condition. These metrics demonstrate that the model with speed layers yields more consistent head rotation velocities within clips, respectively. Furthermore, the "Mean Velocity" metric substantiates that the pre-defined speed value affect the actual velocity level, confirming the efficacy of the speed layers in modulating the synthesized head motion dynamics. In our approach, speech-driven cases are assigned speeds from 0.1 to 1.0, while singing scenarios might use higher settings (1.0-1.3) for faster, song-congruent head motions. Speeds exceeding 1.5 could lead to unnaturally rapid and jittery movements.

\begin{table}[htbp]
  \centering
  \caption{The detected head rotation velocity in the generated videos.}
  \label{tab:speed_layer}
  \begin{tabularx}{\textwidth}{l>{\centering\arraybackslash}X>{\centering\arraybackslash}X>{\centering\arraybackslash}X>{\centering\arraybackslash}X>{\centering\arraybackslash}X}
    \toprule
      & No Speed & 0.1 & 0.7 & 1.3 & 1.9\\
    \midrule
    Mean Velocity & 1.365 & 0.878 & 1.001 &1.162 & 1.246 \\
    Velocity Variance & 1.002 & 0.357 & 0.454 & 0.550 & 0.657 \\
    VMV & 0.134 & 0.046 & 0.054 & 0.057 & 0.058\\
    \bottomrule
  \end{tabularx}
\end{table}

\subsubsection{the Control Effect of the Face Locator.}
\label{exp:facelocator}
\begin{figure}[tb]
  \centering
  \includegraphics[width=\textwidth]{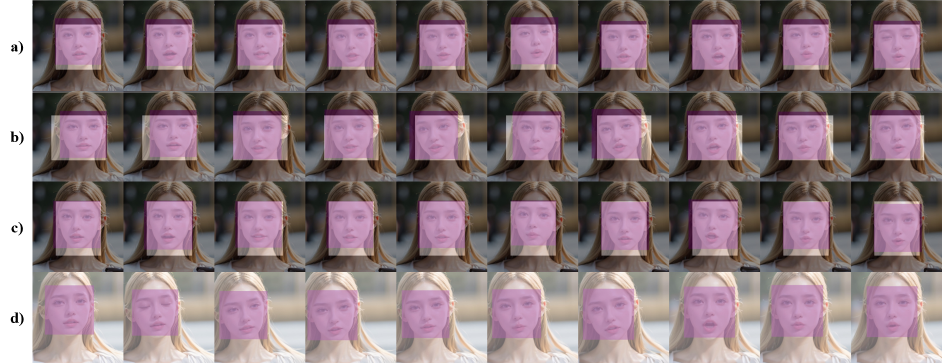}
  \caption{The comparisons on inputting different face regions, the detected facial bboxes in the generated video are denoted by purple regions, while the designated input face regions are represented by white. a) Utilizing the reference image's facial bbox as the face region; b) input an extended bbox with increased width; c) input an expanded bbox with greater height; d) applying a uniform white mask as face region.
  }
  \label{fig:face_loacte}
\end{figure}

Face locator take the face region as input, and delineates the permissible domain for facial movements, thereby influencing the range of head motion. This is illustrated in Figure \ref{fig:face_loacte}, where the character exhibits minimal head movement upon receiving an appropriately sized facial region as input. Conversely, a broader input region permits the character to exhibit more head swings during speech, and an input region with increased height facilitates nodding gestures. Furthermore, inputting a uniform white mask does not provide specific guidance, allowing for facial generation in arbitrary locations. The purple bboxes could extend beyond the prescribed white face region, evidencing that the face locator exerts only a weak condition on head movements, permitting a degree of motion that transcends its indicated boundaries.

\section{Conclusion}

In this work, we introduced EMO, a framework that advances the generation of expressive talking head videos by leveraging audio input to drive the animation process. By eschewing the traditional reliance on intermediate signals, EMO capitalizes on weakly conditioned audio2video diffusion to generate lifelike portraits. Our method demonstrates notable improvements in capturing the nuances of human expressiveness and maintaining consistent identity across video frames. The results outperform existing state-of-the-art methods, confirming the potential of EMO as a powerful tool for creating rich, audio-driven video content.


%
%
\bibliographystyle{splncs04}
\bibliography{main}
\end{document}


\title{EMO: Emote Portrait Alive - Generating Expressive Portrait Videos with Audio2Video Diffusion Model under Weak Conditions (\textit{Supplementary Material})}

\titlerunning{EMO-Emote Portrait Alive}

\author{Linrui Tian\orcidlink{0000-0003-1202-6040} \and
Qi Wang \and
Bang Zhang \and
Liefeng Bo}

\authorrunning{L. Tian, Q. Wang, B. Zhang, and L. Bo}

\institute{Institute for Intelligent Computing, Alibaba Group\\
\email{\{tianlinrui.tlr, wilson.wq, zhangbang.zb, liefeng.bo\}@alibaba-inc.com}\\
\url{https://humanaigc.github.io/emote-portrait-alive/}}
\maketitle

This supplementary material contains additional information that could not be included in the main manuscript due to space limitations. It includes a wider exploration of our method's results, and we discuss about our training dataset. Following that, we provided a candid discussion on the limitations of EMO, and a preview of future research directions.

\section{More results}

\subsection{User study}

\begin{table}[htbp]
  \centering
  \caption{User study.}
  \label{tab:user_study}
  \begin{tabularx}{0.75\textwidth}{l>{\centering\arraybackslash}X>{\centering\arraybackslash}X}
    \toprule
      & lip sync & vividness \\
    \midrule
    Wav2Lip\cite{wav2lip} & 3.90 & 1.66 \\
    SadTalker\cite{sadtalker} & 3.38 & 2.34 \\
    DreamTalk\cite{dreamtalk} & 4.05 & 3.22\\
    MakeItTalk\cite{MakeItTalk} & 1.66 & 2.66 \\
    Ours & \textbf{4.17} & \textbf{4.38} \\
    \bottomrule
  \end{tabularx}
\end{table}

We also carried out a user study using the generated outputs, involving 20 participants, evenly split between 10 males and 10 females, ranging in age from 20 to 60, and with diverse levels of computer technical expertise. For every volunteer, We will show them the results of all methods on the same image and audio simultaneously in each round. The volunteers are asked to rate each video between 1 and 5 (higher is better), in terms of lip synchronization and vividness. The results are shown in the Table \ref{tab:user_study} and our method significantly outperforms other methods, especially on the vividness.



\subsection{Motion frames}
Our method employs motion frames to bolster consistency across clips generated in sequence.
Some other video generation techniques\cite{animate_anyone} may employ 'frame replacing'-substituting the initial $m$ frames with the final $m$ frames from the preceding clip at each denoising step and using temporal modules for frame-to-frame coherence.
Our approach does not utilize strong control signals, such as pose sequence, to guide the sequential generation process of video clips. Thus, along with a sensitivity to 'jump cuts' highlighted in Sec \ref{sec::data}, our method might experience frame discontinuity during clip transitions. 

\subsection{Inference steps}

\begin{figure}[tb]
  \centering
  \includegraphics[width=0.75\textwidth]{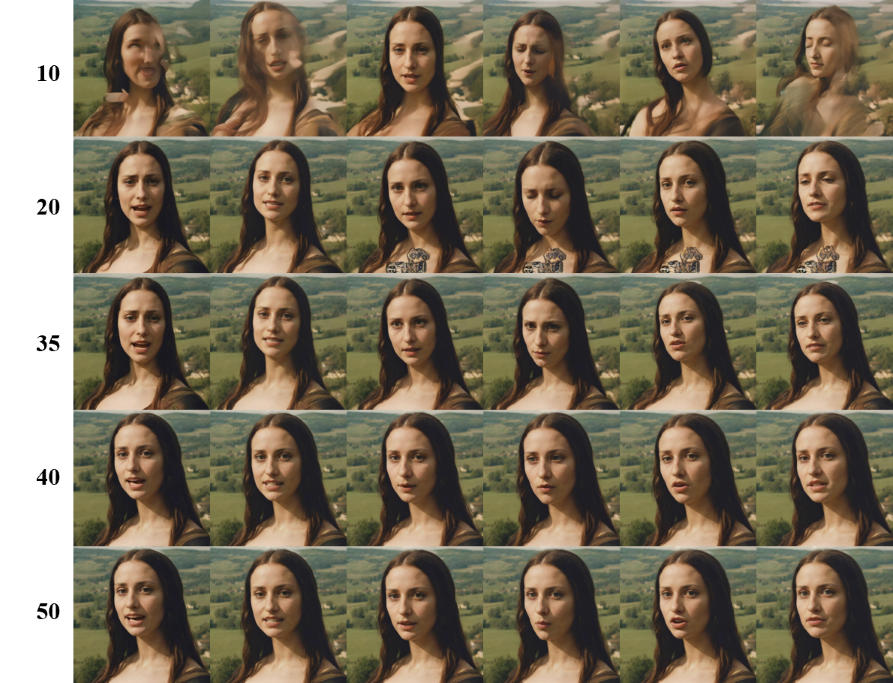}
  \caption{The generated results under different denoising steps.
  }
  \label{fig:steps}
\end{figure}


In our experiments, we observed that the number of denoising steps during inference critically influences the quality of the generated output. As depicted in the Figure \ref{fig:steps}, a suboptimal number of steps (fewer than 20) results in temporal inconsistencies across frames, as well as a prevalence of visual artifacts within the sequences. When the number of denoising steps is marginally increased (within the range of 20–35), the artifacts are somewhat ameliorated, however, the generated characters might exhibit noticeable jitter and instability. Our experiments indicate that employing more than 35 denoising steps enhances stability and temporal coherence, leading to more reliable results in the synthesized sequences.

\section{Dataset}

\subsection{Data overview}

Our training dataset mainly contains 1) HDTF\cite{hdtf}, encompassing 15.8 hours of high-fidelity talking head videos, featuring approximately 362 unique character identities with a majority being anchors and spokesmen; 2) VFHQ\cite{vfhq}, which contains 16k high-resolution talking video clips without audio; 3) A diverse collection of speech and singing data aggregating to 250 hours, sourced from online platforms. 
The videos are publicly accessible, and the collectors are also instructed to carefully review the content to exclude any personally identifiable information, thus ensuring adherence to privacy standards and the terms of use of the platforms. 
Our dataset features over 10,000 unique character identities, with a focus on English and Chinese languages. We ensured that the character's position, camera angle, and background in the collected data remained relatively unchanged, with each frame capturing a single individual. Unlike some talking head methods that rely on Voxceleb\cite{Voxceleb,sun2023vividtalk, dreamtalk}, we opted not to use it due to its frequent centering on facial centroids, leading to unstable camera movements. 

However, as indicated in Table 1 of the main manuscript, our method still exhibits exceptional performance in the absence of our self-collected dataset. Our model yields satisfactory results when trained solely on publicly available datasets. The extensive dataset we compiled primarily enhances facial expressions and video content dynamics.


\subsection{Preprocessing and labeling}

\subsubsection{Preprocessing the orginal videos.}
\label{sec::data}
Discontinuities in training data, such as inconsistencies in character appearance and camera switches, pose significant challenges for model training, often resulting in the generation of unstable videos. To mitigate these effects, our approach involves segmenting videos into shorter clips that maintain both temporal coherence and scene consistency, similar to the method described in SVD\cite{svd}. By employing PySceneDetect for scene transition identification, we ensure that each clip, ranging from 3 to 12 seconds, contributes to a more reliable and stable dataset for training our generative models. Furthermore, 'jump cuts', prevalent in speech videos for maintaining narrative continuity, present additional detection challenges due to their subtle nature. Our solution incorporates 'speed layers' to address the abrupt velocity changes associated with 'jump cuts', thereby enhancing the fluidity and consistency of the generated video content.

\subsubsection{Labeling the data.}
We performed cropping on the video clips based on the expanded facial bounding boxes of the characters within the clips, and converted each clips to 30 FPS. And we label the cropped clips with 1) deploying MediaPipe\cite{mediapipe} to ascertain the facial bounding box in all frames, thereby delineating the facial regions; 2) extracting audio embeddings using the pre-trained Wav2Vec model\cite{wav2vec}; and 3) determining the character's 6-DoF (six degrees of freedom) head pose to calculate frame-by-frame velocities. 

\section{Limitations and future work}

\begin{figure}[tb]
  \centering
  \includegraphics[width=0.75\textwidth]{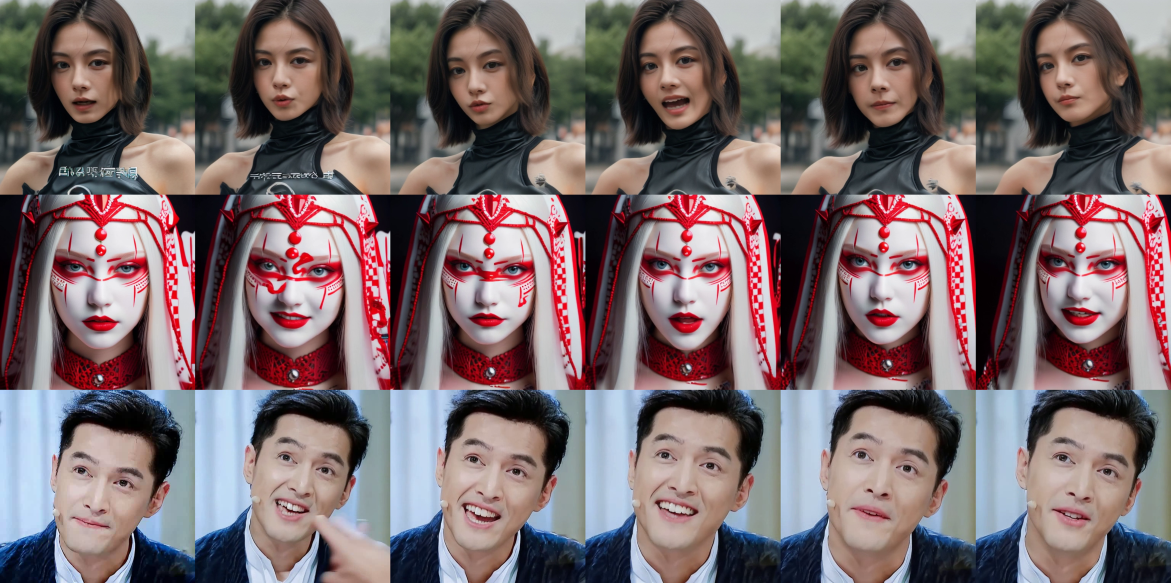}
  \caption{Exhibition of Artifacts: This includes the manifestation of subtitle-like patterns and inaccurately rendered body parts. Notably, in certain instances, the generated frames may self-correct in subsequent sequences.
  }
  \label{fig:artifact}
\end{figure}

As introduced in the main manuscript, our model does not employ explicit control signals to control the character's motion. As shown in the Figure \ref{fig:artifact}, there is a propensity for the model to inadvertently generate body parts into frames, particularly when the audio input features prominently expressive emotions. This tendency arises from the nature of our training dataset, where characters exhibiting pronounced emotional states are frequently associated with more dynamic hand and body movements. Given that our primary focus lies on the head region, the dataset is deficient in data pertaining to other body parts, with a mere 3\% of frames featuring hands. In instances where the model attempts to render these body parts unintentionally, the result is often the generation of incorrect body parts, leading to artifacts.

Additionally, the model may produce caption-like patterns under some circumstances. This issue stems from a subset of the training data sourced from the internet that includes videos with embedded subtitles, hence introducing unwanted textual artifacts into the generated frames. This phenomenon has been observed and is not exclusive to our model; it is also prevalent in the output of Text-to-Image (T2I) models.

To address these issues, one potential solution involves the introduction of control signals for body parts and subtitles. More specifically, utilizing mask-like input, similar to face regions.

Another limitation of our model, EMO, is its reliance on audio as the principal control signal. EMO has been trained to learn the correlation between tonal features in the audio and facial expressions in characters. However, this association can result in expressions for the driven characters that may not always align with user expectations, limiting the ability to produce desired video outcomes in a controlled manner. Introducing a mechanism to define emotions could enhance user convenience by providing more predictable results.

Compared to the other diffusion-free talking head generation methods, EMO is more time-consuming, it generates 12 frames (one clip) per 18 seconds (under 40 denoising steps) on A100 GPU.

Despite these challenges, EMO represents a significant leap forward in the development of highly expressive video generation models, laying a foundation for future innovation in this field. We leave these issues as open questions for subsequent research.

\bibliographystyle{splncs04}
\bibliography{egbib}